\documentclass[conference]{IEEEtran}
\IEEEoverridecommandlockouts
\usepackage{cite}
\usepackage{amsmath,amssymb,amsfonts}
\usepackage{algorithmic}
\usepackage{graphicx}
\usepackage{textcomp}
\usepackage{booktabs}
\usepackage{multirow}
\usepackage{xcolor}
\usepackage{rotating}
\usepackage{subfigure}
\usepackage{colortbl}
\definecolor{mygray}{gray}{.94}

\begin{document}

%\title{Conference Paper Title*\\
\title{Polyp-E: Benchmarking the Robustness of Deep Segmentation Models via Polyp Editing}

\author{\IEEEauthorblockN{
Runpu Wei\IEEEauthorrefmark{1},
Zijin Yin\IEEEauthorrefmark{1},
Kongming Liang\IEEEauthorrefmark{1}, 
Min Min\IEEEauthorrefmark{2},
Chengwei Pan\IEEEauthorrefmark{3}, 
Gang Yu\IEEEauthorrefmark{4}, \\
Haonan Huang\IEEEauthorrefmark{1}, 
Yan Liu\IEEEauthorrefmark{2}, and
Zhanyu Ma\IEEEauthorrefmark{1}} % ,~\IEEEmembership{Fellow,~IEEE}
\IEEEauthorblockA{\IEEEauthorrefmark{1}School of Artificial Intelligence,
Beijing University of Posts and Telecommunications, Beijing, China}
\IEEEauthorblockA{\IEEEauthorrefmark{2}The Fifth Medical Center, Chinese PLA General Hospital, Beijing, China}
\IEEEauthorblockA{\IEEEauthorrefmark{3}School of Artificial Intelligence (Institute of Artificial Intelligence), Beihang University, Beijing, China}
\IEEEauthorblockA{\IEEEauthorrefmark{4}Children's Hospital of Zhejiang University School of Medicine, Zhejiang, China}
\thanks{% Manuscript received December 1, 2012; revised August 26, 2015. 
Corresponding author: Kongming Liang (email: liangkongming@bupt.edu.cn).}}

\maketitle

\begin{abstract}
Automatic polyp segmentation is helpful to assist clinical diagnosis and treatment. In daily clinical practice, clinicians exhibit robustness in identifying polyps with both location and size variations. It is uncertain if deep segmentation models can achieve comparable robustness in automated colonoscopic analysis. To benchmark the model robustness, we focus on evaluating the robustness of segmentation models on the polyps with various attributes (e.g. location and size) and healthy samples. 
Based on the Latent Diffusion Model, we perform attribute editing on real polyps and build a new dataset named Polyp-E. 
Our synthetic dataset boasts exceptional realism, to the extent that clinical experts find it challenging to discern them from real data. We evaluate several existing polyp segmentation models on the proposed benchmark. The results reveal most of the models are highly sensitive to attribute variations. As a novel data augmentation technique, the proposed editing pipeline can improve both in-distribution and out-of-distribution generalization ability. 
The code and datasets will be released.
\end{abstract}


% \fi
\IEEEpeerreviewmaketitle

\section{Introduction}
Colorectal Cancer (CRC) is the third most commonly diagnosed cancer \cite{siegel2020colorectal} in the world. As the gold standard for CRC screening and prevention, colonoscopy plays a crucial role in identifying adenomatous polyps and reducing mortality rates. Clinicians exhibit robustness in identifying polyp locations and size variations \cite{ali2024assessing}, but it's uncertain if the deep learning model exhibits similar robustness in automated colonoscopic analysis due to real-world physical limitations. As these technologies become more widespread \cite{ref_unet,ref_pranet,ref_uacanet,ref_sanet,ref_msnet,ref_m2snet,ref_cfanet,ref_ccldnet,xiao2022icbnet,ref_polyppvt,huang2022transmixer,ref_endofm,ref_polypsam,ref_medsam,ref_ppsam,xia2024dusformer}, issues of robustness are becoming increasingly crucial. Especially when faced with various clinical scenes, biomedical visual models may become unhelpful or lead to missed diagnoses. Thus, it's crucial to assess their biases and failures accurately before practical deployment. 

Previous researchers have provided large-scale polyp segmentation datasets \cite{ref_sunseg,ref_piccolo,ref_kvasir}, but they do not investigate models' sensitivity under polyp attributes fine-grained changes. Furthermore, since the lack of sufficient samples with polyp location and size changes in real datasets, it is difficult to distinguish whether degeneration is caused by attribute variations or other information changes. Recently, diffusion models \cite{ref_ddpm,ref_ddim,ref_ldm} show promise in achieving realistic medical images generation \cite{ref_sdm,ref_arsdm} and object attribute editing \cite{ref_imagenete}. However, existing approaches still suffer from unstable image quality, background perturbation, and fuzzy object boundaries. Therefore, we focus on editing the polyp attributes and evaluating segmentation model robustness, aiming to provide new insights for enhancing diagnostic accuracy. 

\begin{figure}[t]
\centerline{\includegraphics[width=\linewidth]{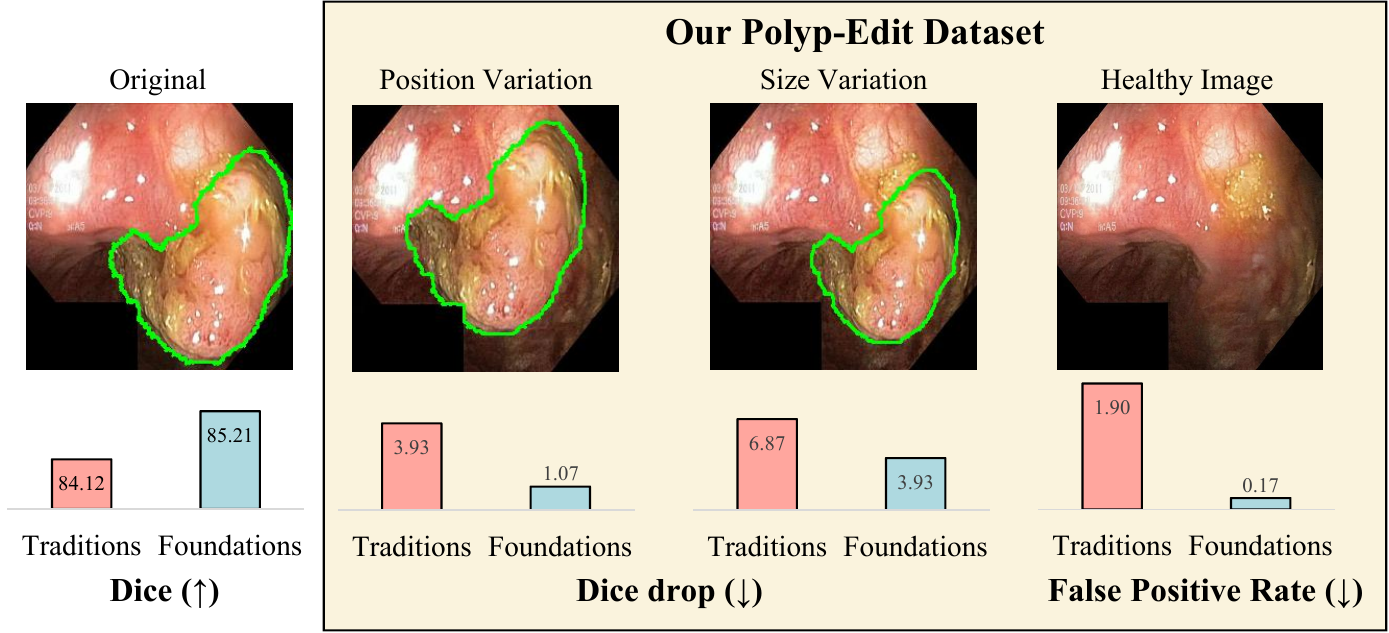}}
\caption{An overview of our study. We generate healthy, size and position variations utilizing real polyp images, and then evaluate the robustness of various state-of-the-art segmentation models including traditional models and foundation models.}
\label{fig_intro}
\end{figure}

\begin{figure*}[ht!]
    \centering
    \includegraphics[width=\linewidth]{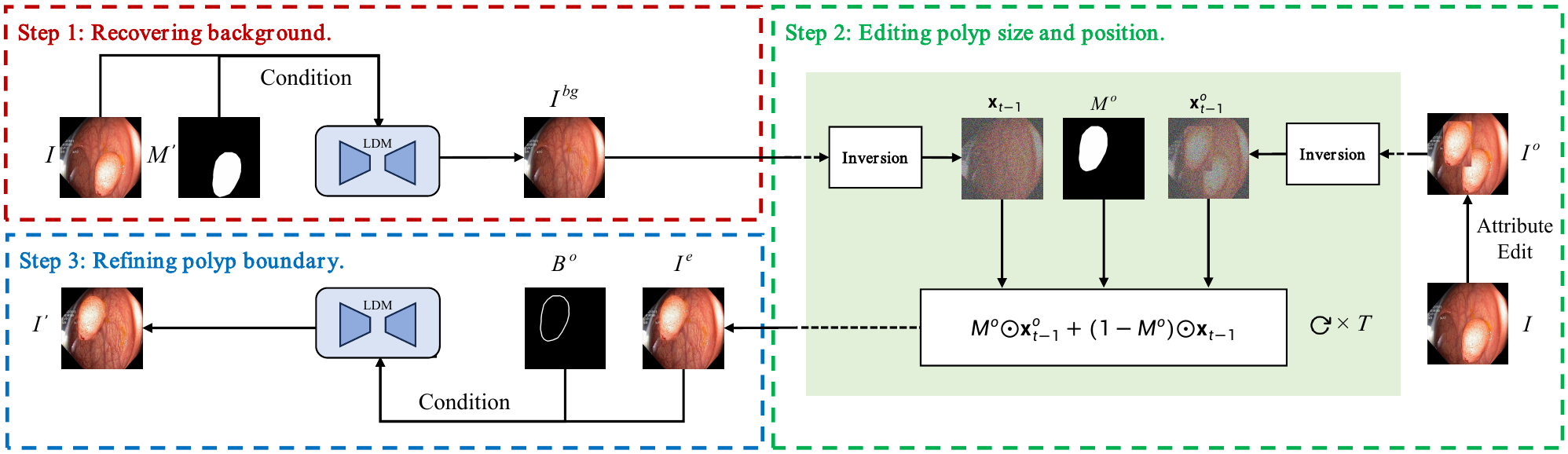}
    \caption{The illustration of our polyp editing pipeline. (1) We disentangle the background and polyp, then in-paint the missing part of the background. This process yields a healthy tissue sample that corresponds to the original polyp image, effectively separating the pathological area from its surroundings while maintaining the context.
    (2) We manipulate real images through the precise adjustment of polyp size and position, thereby customizing their appearance and layout.
    (3) We refine boundaries utilizing in-painting to generate more faithful samples.}
    \label{fig_pipeline}
\end{figure*}
%Specifically, we investigate the application of deep generative models for assessing the robustness of polyp segmentation models. 
In this paper, we investigate benchmarking and improving the robustness of polyp segmentation methods based on the popular generation model Latent Diffusion Model \cite{ref_ldm}.
As shown in Fig. \ref{fig_intro}, we perform editing on real polyp and consider three data editing scenarios including healthy (non-polyp), size variations, and position variations, we finally construct a new benchmark \textbf{Polyp-E(dit)}, which contains samples that can occur in the real world but don't appear or are underrepresented in the real datasets due to physical limitations. We only minimally modify the images. Hence, it better retains fidelity and diversity. We benchmark various polyp segmentation models such as U-Net\cite{ref_unet}, PraNet\cite{ref_pranet}, Endo-FM\cite{ref_endofm}, observe that most models have varying degrees of sensitivity to various attribute changes, and existing models have a certain degree of over-segmentation for healthy samples. We suggest that considering healthy samples can avoid over-segmentation and improve model generalization ability.

In summary, our contributions are: 
\begin{enumerate}
    \item A novel approach is proposed to edit polyp images via attribute manipulation.
    \item We construct the Polyp-E benchmark to evaluate model robustness towards different attributes. It opens up new avenues to study the robustness of medical image analysis against lesion attributes.
    \item We conduct extensive experiments and find that polyp segmentation models exhibit varying sensitivity to attribute variations.
    \item As a novel data augmentation technique, our method can improve both in-distribution and out-of-distribution generalization ability.
\end{enumerate}

\section{Related works and Preliminary}
\subsection{Polyp Segmentation}
Colon polyps are a deserved precursor to colon cancer, thus accurate segmentation of polyps can reduce misdiagnosis of colon cancer \cite{liao2020exploration}. Various methods have been proposed to achieve effective polyp segmentation.

CNN is one of the most widely used neural network architectures in polyp segmentation. A typical example is U-Net \cite{ref_unet}, a classic segmentation network. PraNet \cite{ref_pranet} integrates the Reverse Attention module \cite{chen2018reverse}, which highlights the boundaries between polyps and their surroundings. UACANet \cite{ref_uacanet} takes into account an uncertain area of the saliency map, SANet \cite{ref_sanet} designs a color exchange operation to decouple the image contents and colors, prompting the model to focus more on the target shape and structure. MSNet \cite{ref_msnet} introduces a multi-scale subtraction network to avoid generating redundant information, which can lead to inaccurate localization and blurred polyps edges.
Another architecture gaining traction in polyp segmentation is Transformers \cite{min2022transformer}, which enables the model to understand deeper contextual dependencies effectively. A notable example should be Polyp-PVT \cite{ref_polyppvt}.
Recently, there has been a growing interest in using SAM \cite{ref_sam} for polyp segmentation, which can perform well with limited data. Noteworthy implementations include Polyp-SAM \cite{ref_polypsam} and MedSAM \cite{ref_medsam}.

\subsection{Robustness evaluation}
In recent years, benchmarking and improving model robustness have attracted more and more attention. ImageNet-C \cite{hendrycks2019benchmarking} introduced a robustness benchmark to assess the impact of common perturbations and corruptions on ImageNet. ImageNet-A \cite{hendrycks2019using} gathered natural adversarial examples from other unlabeled datasets. These benchmarks are effective for identifying model weaknesses and finding suggestions for improving robustness. Attribute editing dataset creation is a new area of research with few studies exploring it. Among them, ImageNet-E \cite{ref_imagenete} stands out as a recent benchmark specifically designed to evaluate image classifier robustness of object attributes including background settings, object sizes, spatial positioning, and orientation.

However, it is important to note that not all improvements in general computer vision benchmarks translate well to medical image analysis. Medical images differ from natural images, the general robustness benchmarks may not fully cover the specific challenges of medical image analysis. 

In the broader scope of medical image analysis, recent research has highlighted the importance of assessing the robustness of deep neural networks against lower image quality. 
Previous endoscopic computer vision challenges \cite{ali2022endoscopic} have addressed the issue of diminished image quality and generalization across different endoscopic modalities.
Some researchers have proposed stress testing \cite{eche2021toward}, a form of robustness evaluation, as a strategy to address underspecification in radiology. The authors suggest that robustness evaluation can be designed by modifying medical images or selecting specific testing datasets.
Since then, ROOD-MRI \cite{boone2023rood} conducted a study on segmentation models using MRI data that are out-of-distribution and corrupted. They concluded that modern CNN architectures are highly susceptible to distribution shifts, corruptions, and artifacts.
Similar comparable results have been shown in skin disease classification task \cite{maron2021benchmark, young2021stress}, skin cancer-classification models that meet conventional metrics require further validation through computational stress tests to assess clinical readiness.
\cite{islam2023robustness} evaluated the robustness performance of both skin cancer classification and chest X-ray scans, they suggest robustness evaluation should be a standard practice in clinically validating image-based disease detection models.

\subsection{Preliminary}
To contextualize our editing pipeline, which hinges on latent diffusion models, we will recapitulate the essentials of the diffusion process \cite{ref_ddim,ref_ddpm}, which elucidate their role in generating images, serving as a critical foundation for comprehending our methodology.

Inspired by non-equilibrium thermodynamics \cite{ref_nonequilibrium}, diffusion models \cite{ref_ddim,ref_ddpm} are probabilistic generative models designed to learn a data distribution $p\left (x\right )$ by denoising a normally distributed variable gradually, which corresponds to learning the reverse process of a fixed Markov Chain of length $T$.
For image synthesis, it can be interpreted as an equally weighted sequence of denoising autoencoders $\epsilon_\theta \left(x_t,t\right)$; $t=1\ldots T$, which are trained to predict a denoised variant of their input $x_t$, where $x_t$ is a noisy version of the input $x$. The corresponding objective can be simplified as:

\begin{equation}
    L_{DM}=\mathbb{E}_{x,\epsilon \sim \mathcal{N} \left( 0,1\right),t} \left [|| \epsilon - \epsilon_\theta \left ( x_t,t \right ) ||^2_2) \right],
\end{equation}
with $t$ uniformly sampled from $\left \{ 1,\ldots,T \right \}$.

Diffusion models achieve expeditious evolution in image synthesis \cite{ref_ddpm,ref_ddim,ref_ldm,ref_imagenete,ref_arsdm}. Compared to the image space, latent space is more suitable for likelihood-based generative models. In this way, we can pay more attention to the important semantic bits of data and increase computational efficiency. Thus, we leverage the state-of-the-art image generation algorithm Latent Diffusion Model (LDM) \cite{ref_ldm}, in which the DDIM diffusion process \cite{ref_ddim} performs in low-dimensional latent space where semantic information can better transfer. It consists of a variational autoencoder network \cite{ref_vae} to encode and decode between latent space and pixel space, and denoising network U-Net \cite{ref_unet} architecture conditioned on the guiding input prompt to achieve diffusion process. 

%\cite{ref_ddpm,ref_ddim} are probabilistic models designed to learn a data distribution $p\left (x\right )$ by denoising a normally distributed variable gradually, which corresponds to learning the reverse process of a fixed Markov Chain of length $T$. For image synthesis, it can be interpreted as an equally weighted sequence of denoising autoencoders $\epsilon_\theta \left(x_t,t\right)$; $t=1\ldots T$, which are trained to predict a denoised variant of their input $x_t$, where $x_t$ is a noisy version of the input $x$. The corresponding objective can be simplified as:
%\begin{equation}
%    L_{DM}=\mathbb{E}_{x,\epsilon \sim \mathcal{N} \left( 0,1\right),t} \left [|| \epsilon - \epsilon_\theta \left ( x_t,t \right ) ||^2_2) \right],
%\end{equation}
%with $t$ uniformly sampled from $\left \{ 1,\ldots,T \right \}$.

%\noindent \textbf{Latent Diffusion Model(LDM). } As ~\cite{ref_ldm} proposed, we can use the variational autoencoder (VAE) ~\cite{ref_vae} to encode an image $\mathbf{X}$ into VQGAN latent space and obtain the latent variable $\mathbf{x}$ corresponding to the image $\mathbf{X}$. Compared to the image space, latent space is more suitable for likelihood-based generative models. In this way, we can pay more attention to the important semantic bits of data and increase computational efficiency.

\section{Methods}
In this section, we provide a detailed explanation of our polyp attribute editing pipeline.

\subsection{Polyp-E Dataset Construction}
\label{sec_editmethod}
Our objective is to construct different polyp attribute variations using the diffusion-based pipeline. The overview of our proposed pipeline is illustrated in Fig. \ref{fig_pipeline}. The details of each component are described as follows.

\subsubsection{Recovering background.}

\begin{figure}[]
    \centering
    \includegraphics[width=\linewidth]{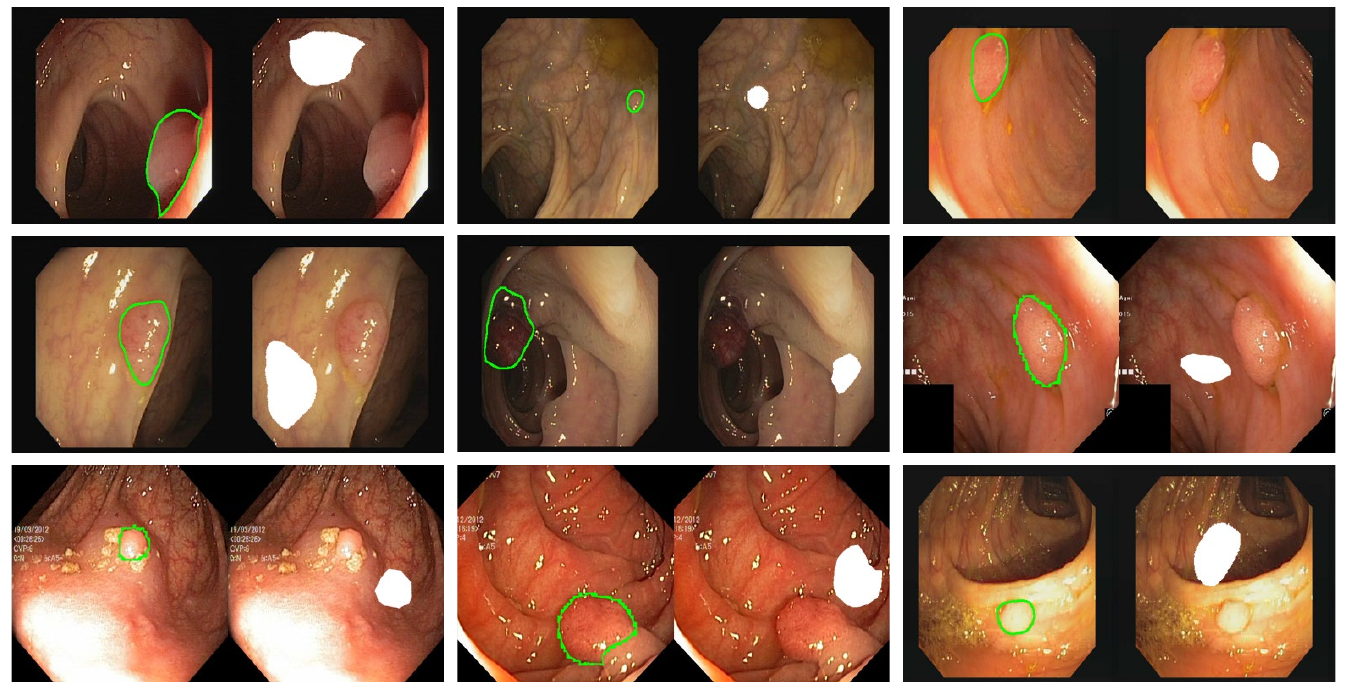}
    \caption{The visualization of the random mask generation strategy of the recovering background part.
    Each sub-figure (Left: Original image. Right: Corresponding masked image with random mask.) is an example of a pair of images used for training polyp to non-polyp image transformation.}
    \label{fig_random_mask}
\end{figure}

%Thus, we should recover the background information of a given polyp image, transforming it into a corresponding healthy image. 
%Since diffusion models iteratively refine an image starting from random noise, they are easily adapted for inpainting tasks when the image and mask are given as input, so we focus on using diffusion models to accomplish this goal.
To ensure a valid evaluation across different attributes, all other variables must remain constant, particularly the background. 
Therefore, we first recover corresponding background information for a given polyp image by in-painting. To achieve this, we proposed a training strategy. Specifically, for a given original image $I$ and its polyp mask $M$, we first generate a mask $M'$ randomly covering a part of the background of $I$ as shown in Fig. \ref{fig_random_mask}. To create the random mask $M'$, we apply fundamental image augmentation including rotation, scaling, etc to its corresponding polyp mask $M$. Since the network focuses on reconstructing the non-polyp image portion, the generated masks are randomly placed in the original images, avoiding overlapping with the polyp portion.

Then we take the concatenation of $I$ and $M'$ as the input condition, the original image $I$ serves as the target to train the diffusion model. In this way, we can obtain an in-painting diffusion model capable of exclusively reconstructing the background.

\begin{figure}[!h]
    \centering
    \includegraphics[width=\linewidth]{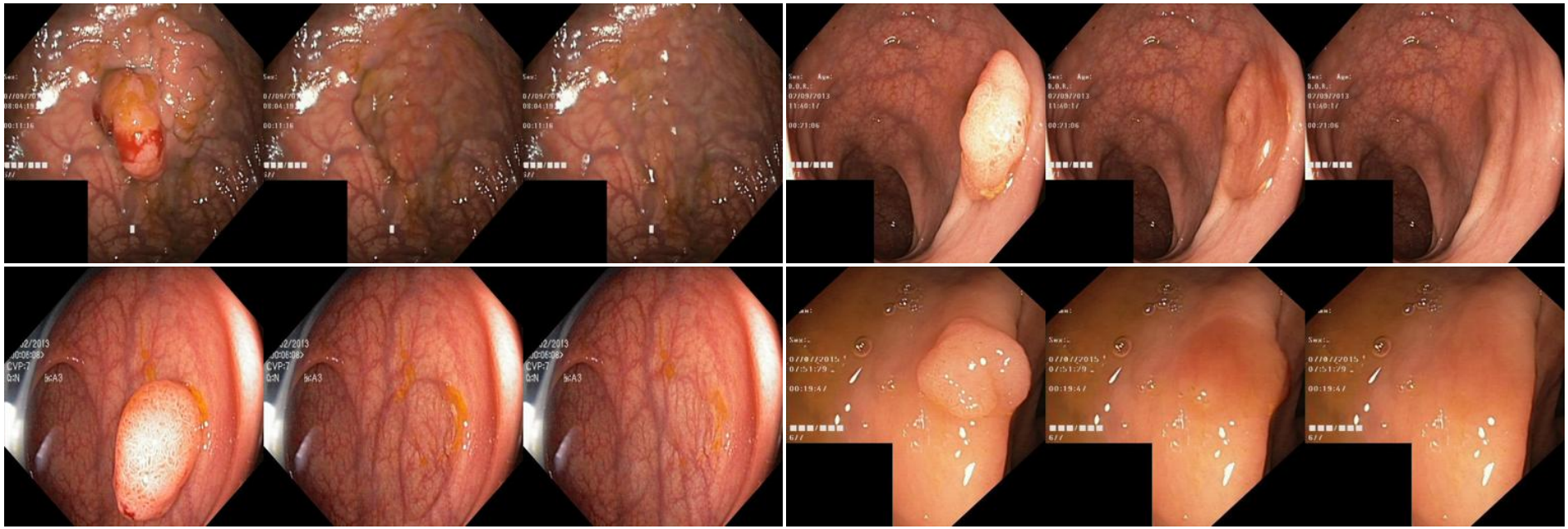}
    \caption{Example of negative images inpainted by different masks in inference phase. For each subfigure, from left to right are the original polyp image, the negative image in-painted with the original polyp mask and the negative image in-painted with the 20-pixel dilated polyp mask.}
    \label{fig_dilate}
\end{figure}

In the inference phase, the combination of the original image $I$ and its polyp mask $M$ serves as an input condition to generate background sample $I^{bg}$. 
However, we find there are obvious artifacts of polyp boundary in $I^{bg}$, shadows or connections around the polyp will remain as shown in Fig. ~\ref{fig_dilate} if using the original polyp mask. These lingering shadows manifest as unwanted artifacts, undermining the authenticity of the negative image being generated. To address this issue, we dilate the original polyp mask $M$ by $20$ pixels as the final mask for the input condition in the inference phase.

\subsubsection{Editing polyp attributes.}
To control polyp attributes such as size and position, we first apply the affine transformation to the specific polyp area, and then we seamlessly integrate the transformed polyp into its corresponding background utilizing the diffusion process.

For attributes controlling, let $E$ denote the affine transformation operation and $P$ represent the pixel locations of the polyp in the image $I$, where $P \in \mathbb{R}^{N_o \times 3}$. $N_o$ stands for the number of pixels belonging to the polyp, and $p_i = {\left[x_i, y_i, 1\right]}^{\mathsf{T}}$ denotes the position of the $i$-th pixel of the polyp. Then the newly transformed polyp image is $I^{o}=I$ and $I^{o}[E\cdot P] = I[P]$, and the transformed polyp mask $M^{o}=0$ and $M^{o}[E \cdot P] = M[P]$. The transformation matrix $E$ for size and position translation is as follows respectively:

\begin{equation}
    E_{\text {size}}=\left[\begin{array}{ccc}s & 0 & \Delta x \\0 & s & \Delta y \\0 & 0 & 1\end{array}\right],
    E_{\text {position}}=\left[\begin{array}{lll}1 & 0 & w^{\prime} \\0 & 1 & h^{\prime} \\0 & 0 & 1\end{array}\right],
\label{equ_attributeEdit}
\end{equation}

where $s$ denotes the amplitude of size change. $\Delta x=\left(1-s\right) \cdot \left(x+w/2\right), \Delta y=\left(1-s\right) \cdot \left(y+h/2\right)$, $w^{\prime}$ and $h^{\prime}$ represent the amount of translation in the horizontal and vertical directions respectively, and $[x,y,w,h]$ represents the enclosing rectangle of the corresponding mask $M$. We use $\tau$ to denote the amplitude of position translation, where $\tau=w^\prime/w=h^\prime/h$.

We leverage an unconditional Latent Diffusion Model to blend the transformed polyp image and the corresponding background at different noise levels along the diffusion process. 
We denote $\mathbf{x}^{o}$ as the latent embedding of transformed polyp image $I^{o}$, and $\mathbf{x}^{bg}$ as the latent embedding of recovered background $I^{bg}$ at initial time step $t_{0}$.
At each time step $t$, we perform a guided diffusion step with a latent $\mathbf{x}_{t}$ to obtain the $\mathbf{x}_{t-1}$. At the same time, we obtain a noised version of attribute-edited latent polyp $\mathbf{x}^{o}_{t-1}$. 
Then the two latent are blended under the constraint of mask $M^{o}$ as:

\begin{equation}
    \mathbf{x}_{t-1}=M^{o} \odot \mathbf{x}^{o}_{t-1} + (1-M^{o}) \odot \mathbf{x}_{t-1}
\label{equ_blend}
\end{equation}

At last, we can obtain the edited image $I^{e}$ with polyp attribute variation.

\subsubsection{Refining polyp boundary.}
To prevent segmentation models from collapsing under irrelevant pixel perturbation in generated images, \textit{e.g.} adversarial attack \cite{ref_attack}, we consider potential artifacts and errors, especially the boundary area. Therefore, we refine the polyp boundary in $I^{e}$ by re-painting the boundary area.
Specifically, we adopt the combination of images and their boundary masks derived from the original masks as the condition to train a Latent Diffusion Model.
This enables the model to achieve a more natural repainting between the polyp and the background. 
Given the edited image $I^{e}$ and its boundary mask $B^{o}$, we obtain the final refined sample $I^{\prime}$ with more credibility and fidelity.

\section{Polyp-E dataset.}
\begin{figure*}[!h]
    \centering
    \includegraphics[width=\linewidth]{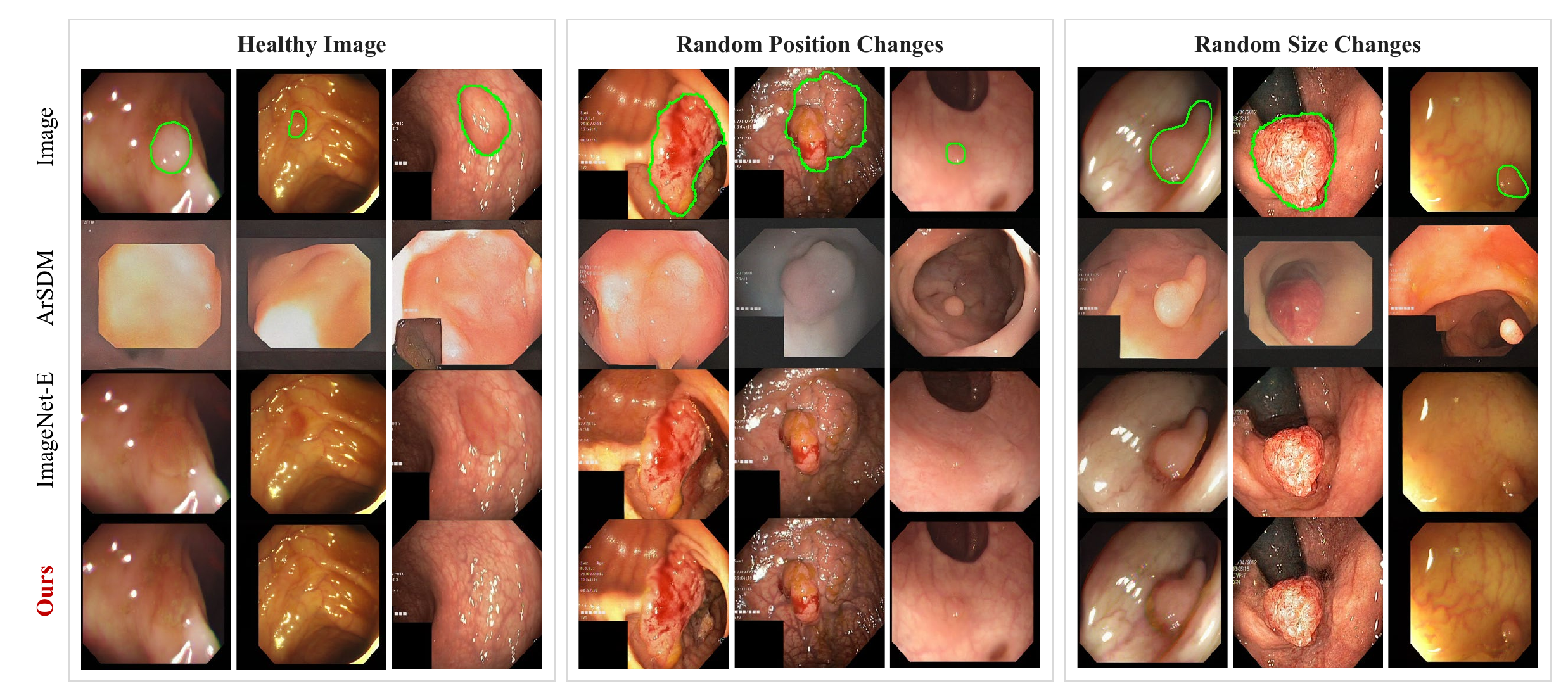}
    \caption{Qualitative comparison of images generated by ArSDM, ImageNet-E, and the proposed method. From top to bottom corresponding to the healthy image, random position changes within 20\% range, and random size changes within 20\% range.}
    \label{fig_vasualization}
\end{figure*}

Using the proposed pipeline above, we construct the Polyp-E dataset based on a total of $798$ test images adopted in \cite{ref_pranet}, which consists of five public polyp segmentation datasets CVC-300 \cite{ref_endoscene}, CVC-ClincDB \cite{ref_clinicdb}, CVC-ColonDB \cite{ref_colondb}, ETIS \cite{ref_etis} and Kvasir \cite{ref_kvasir}. 
The Polyp-E dataset has $3$ components: healthy (pure background) images, images with polyp sizes randomly resized within the ranges of $s=[-0.1,0.1]$, $[-0.2,0.2]$, and $[-0.3,0.3]$, and images with polyps position randomly moved within the ranges of $\tau=[-0.1,0.1]$ and $[-0.2,0.2]$. Qualitative examples can be seen in Fig.\ref{fig_vasualization}.
Clinical experts participate in the manual filtering of the generated samples to further improve the dataset reliability.
Finally, the curated benchmark consists of $4417$ images ($761$ healthy images, $770$ size variations within 10\% range, $742$ size variations within 20\% range, $714$ size variations within 30\% range, $737$ position variations within 10\% range, and $693$ size changes within 20\% range). % 761+737+693+770+742+714

%\subsection{Implementation details.}
For the Latent Diffusion Model, we adopt the pre-trained VAE provided by \cite{ref_ldm}, and train the diffusion model from scratch. 
Following the experimental setting in PraNet \cite{ref_pranet}, 900 pairs from Kvasir \cite{ref_kvasir} and 550 pairs from CVC-ClinicDB \cite{ref_clinicdb} for a total of 1450 image-mask pairs are taken as the training set for the diffusion model. The training images are resized to $512\times512$, and training masks are resized to $128\times128$. 
We set the diffusion iteration timesteps to 50, 20, and 20 in each editing step, respectively. We use a single A800 GPU to conduct all experiments.

% Specifically, we first perform polyp editing using the methods described in the \ref{sec_negative_generation} section, then gather all edited samples and subject them to filtration by expert medical professionals to assemble the proposed dataset. In our experiment across three stages, we set the iteration steps $t_0$ as 50, 20, and 20 steps respectively in Equ~\ref{equ_2}, and it further decreases the number of updates used for our edits. This allows us to edit images in $\sim 5$ seconds on a single A800 GPU.
% These synthetic test sets are used to evaluate the robustness of the polyp segmentation model by comparing performance to the real test set. A significant drop in performance indicates that the segmentation model is not robust to the synthetic dataset on different attributes. 

\section{Experiments}
In this section, we first assess the quality of our edited images. Then, we benchmark various segmentation models on our curated benchmark Polyp-E. Finally, we explore the applications of our proposed pipeline in improving segmentation robustness.

\subsection{Image Quality Assessment}
Our primary concern is the reliability of edited images since our proposed pipeline serves as an evaluator for segmentation models. To assess the quality of our generated images, we compare our pipeline with the popular diffusion-based image generation approach ArSDM \cite{ref_arsdm} and the editing toolkit in ImageNet-E \cite{ref_imagenete}. We use their default settings and the same transformed polyp masks as input to synthesize images. We utilize the \textit{Fréchet Inception Distance (FID)} \cite{ref_fid} and the traditional metric \textit{MS-SSIM}, which measures the distributional and structural similarity between real and generated images, respectively. 

Furthermore, we conduct an expert review to measure the votes of radiologists (with 5 years of experience) for real and fake images using part of each sample (45 images for the real set and 135 images for each synthetic set). It is worth mentioning that we do not adopt expert filtering to ensure the fairness of quality assessment.

\begin{table}
    \centering
    \footnotesize
    \renewcommand{\arraystretch}{1.3} %行距
    \setlength{\tabcolsep}{4pt} % 像素列宽，默认6像素
    \caption{The quantitative results of the image quality assessment.}
    \label{tab_image_quality_assessment}
    \begin{tabular}{lcccc}
    \toprule
    \multirow{2}{*}{}      & \multicolumn{2}{c}{Expert Review} & \multirow{2}{*}{FID ($\downarrow$)} & \multirow{2}{*}{MS-SSIM ($\uparrow$)} \\ \cline{2-3}
                                    & Real ($\uparrow$)       & Fake ($\downarrow$)      &       &      \\ \hline
    Real                            & 97.8\%(44)  & 2.2\%(1)   & 0     & 100    \\
    ImageNet-E \cite{ref_imagenete} & 83.7\%(113) & 16.3\%(22) & 50.3  & 64.6 \\
    ArSDM \cite{ref_arsdm}          & 81.5\%(110) & 18.5\%(25) & 111.5 & 34.4 \\
    \rowcolor{mygray} \textbf{Polyp-E} & \textbf{94.8\%(128)} & \textbf{5.2\%(7)} & \textbf{28.2}        & \textbf{92.4}                     \\ 
    \bottomrule
    \end{tabular}
\end{table}

\begin{figure*}[!t]
    \centering
    \includegraphics[width=0.8\linewidth]{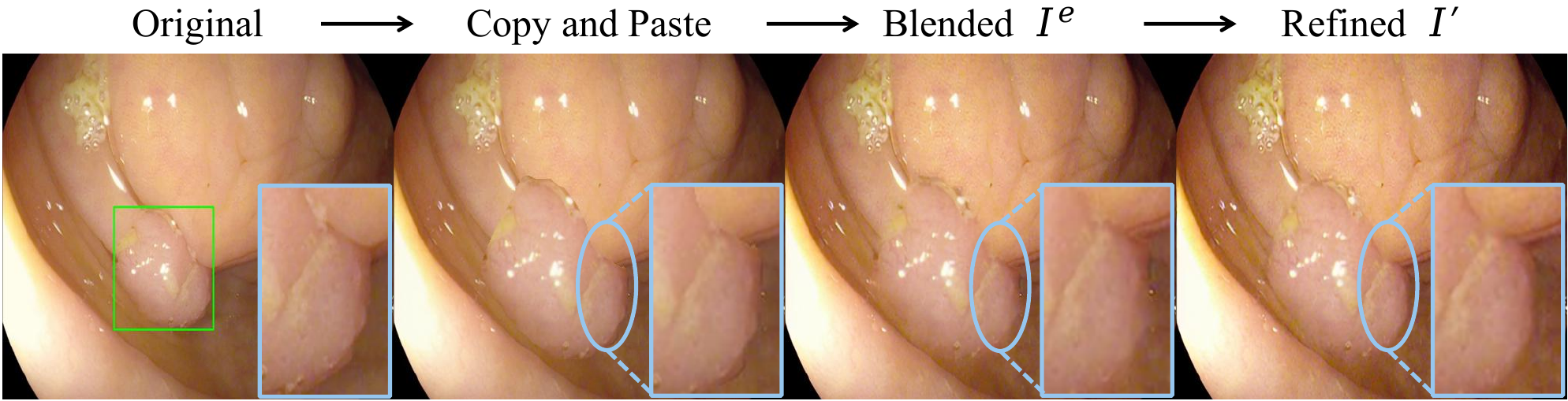}
    \caption{Qualitative results of editing polyp size using our pipeline. It is obvious that the diffusion model blends foreground and background well, and our boundary refinement has better object contours and intricate details.}
    \label{fig_ablation}
\end{figure*}

In Tab. \ref{tab_image_quality_assessment}, both human study and automatic evaluation indicate that our proposed pipeline achieves the best distributional and structural proximity to the real dataset. To more intuitively demonstrate the effectiveness of each component in our pipeline, we conduct the visualization comparison as shown in Fig. ~\ref{fig_ablation}.

Moreover, to avoid the influence of noise inevitably introduced by the diffusion process, we explore the impact of the diffusion process itself on segmentation performance. We reconstruct images using our pipeline, this operation first adds noises to the original images and then denoises them without any attribute transformation. The experimental results can be found in the “Recon.” column in Tab. \ref{tab_robustness} which the performance degradation is negligible compared to degradation induced by attribute changes. At last, we verify the effectiveness of each step of our pipeline in Fig. \ref{fig_ablation}. The proposed pipeline exhibits a remarkable augmentation in both polyp image credibility and fidelity.

% Of the 450 presented images, 97.78\% of real images were marked as real by clinical experts. Images edited via the Polyp-E method attained a 94.81\% authenticity rating from clinical practitioners. Conversely, images edited via Imagenet-E and generated by ArSDM yielded authenticity ratings of only 83.70\% and 81.48\% respectively.
% The results of the expert study are shown in Tab.~\ref{tab_expert_study}. Designating all sets except Real as the “Fake” set of images, we obtain an accuracy, precision, and recall of xxx\%, xxx\%, and xxx\% respectively. Moreover, this study confirms that the generations are indeed good-quality data points, although some may identified as synthetic.
% \subsubsection{Automatic Evaluation.}
% TODO
% For our expert review of the generated image, we had a radiologist (with 5 years of experience) review a total of 350 chaotic images above; The FID ~\cite{ref_fid} scores measure the distance between the output images and the dataset, as well as the variance of the output images. ~\cite{ref_fid2} showed that differences in image compression and resizing, among other factors may induce significant variation in the FID scores. Hence, we used the method introduced in ~\cite{ref_fid2} to compute the FID.

\subsection{Robustness Evaluation}
%\cw{What is the experimental setting? For a trained model, use different augmented images to evaluate the robustness at the inference stage?}
Using our Polyp-E at inference, we benchmark various polyp segmentation architectures including convolution-based models \cite{ref_unet,ref_pranet,ref_uacanet,ref_sanet,ref_msnet,ref_m2snet,ref_cfanet}, transformer-based model CCLDNet \cite{ref_ccldnet} and Polyp-PVT \cite{ref_polyppvt}, and recent foundation model Endo-FM \cite{ref_endofm} and SAM-based model \cite{ref_polypsam, ref_medsam} using their default settings and weights. For synthetic healthy (non-polyp) samples, we use the \textit{False Positive Rate (FPR)}, which refers to the ratio of negative pixels that are incorrectly predicted as positive by the model. For polyp samples, we employ the \textit{Dice} to measure segmentation performance, and the \textit{Dice} drop to measure attribute robustness. For quantitative results of the SAM-based model, we avoid comparing it with other segmentation methods, focusing instead on the SAM family (SAM-B, SAM-L, and MedSAM).

% We conduct evaluation experiments on various architectures including both CNNs (Unet~\cite{ref_unet}, Pranet~\cite{ref_pranet}, UACANet~\cite{ref_uacanet}, SANet~\cite{ref_sanet}, MSNet~\cite{ref_msnet}, M$^2$SNet~\cite{ref_m2snet}, CFANet~\cite{ref_cfanet}), transformer-based models (CCLDNet~\cite{ref_ccldnet}, Polyp-PVT~\cite{ref_polyppvt}), and recently proposed segment anything model (Polyp-SAM~\cite{ref_polypsam}, MedSAM~\cite{ref_medsam}, Endo-FM~\cite{ref_endofm}) with default settings, and evaluated them using mean Dice metrics. Dice measures the accuracy of the model by calculating the overlap between the predictions and the ground truth.

\begin{table*}[!h]
\centering
\renewcommand{\arraystretch}{1.2} % 行距
\setlength{\tabcolsep}{4pt} % 像素列宽，默认6像素
\caption{Quantitative results of polyp segmentation models under difference attribute variations in our Polyp-E dataset. The best results of traditional methods and foundation models are separately bold.}
\label{tab_robustness}
\resizebox{\linewidth}{!}{
\begin{tabular}{crc|c|c|c|cccccc}
\toprule
& & \multirow{3}{*}{Year} & \textit{Dice} ($\uparrow$) & \textit{FPR} ($\downarrow$) & \multicolumn{7}{c}{\textit{Dice drop} ($\downarrow$)} \\ \cline{4-12} 
& \multicolumn{1}{c}{} &  & \multirow{2}{*}{Real} & \multirow{2}{*}{Health} & \multicolumn{1}{c|}{\multirow{2}{*}{Recon.}} & \multicolumn{3}{c|}{Size} & \multicolumn{2}{c|}{Position} & \multirow{2}{*}{Avg.} \\
& \multicolumn{1}{c}{} &  &  & \multicolumn{1}{c|}{} &  & 0.1 & 0.2 & \multicolumn{1}{c|}{0.3} & 0.1 & \multicolumn{1}{c|}{0.2} &  \\ 
\hline \hline

\multirow{11}{*}{\begin{sideways}Traditions\end{sideways}} & U-Net \cite{ref_unet} & 2015 & \multicolumn{1}{c|}{68.91} & \textbf{0.74} & 8.15 & 15.45 & 16.62 & \multicolumn{1}{c|}{18.18} & 17.61 & \multicolumn{1}{c|}{20.60} & 16.10 \\
& PraNet \cite{ref_pranet} & 2020 & \multicolumn{1}{c|}{74.58} & 3.34 & 6.37 & 12.39 & 11.88 & \multicolumn{1}{c|}{14.29} & 14.79 & \multicolumn{1}{c|}{17.30} & 12.84 \\
& UACANet-L \cite{ref_uacanet} & 2021 & \multicolumn{1}{c|}{80.28} & 1.78 & \textbf{3.90} & 10.19 & 11.87 & \multicolumn{1}{c|}{12.42} & 11.49 & \multicolumn{1}{c|}{14.41} & 10.71 \\
& UACANet-S \cite{ref_uacanet} & 2021 & \multicolumn{1}{c|}{79.82} & 2.79 & 4.84 & 9.60 & 10.71 & \multicolumn{1}{c|}{11.91} & 10.46 & \multicolumn{1}{c|}{12.91} & 10.07 \\
& SANet \cite{ref_sanet} & 2021 & \multicolumn{1}{c|}{80.15} & 2.59 & 4.72 & 8.55 & 8.96 & \multicolumn{1}{c|}{10.12} & \textbf{9.10} & \multicolumn{1}{c|}{11.78} & 8.87 \\
& MSNet \cite{ref_msnet} & 2021 & \multicolumn{1}{c|}{79.21} & 3.47 & 6.17 & 11.96 & 12.81 & \multicolumn{1}{c|}{13.64} & 13.16 & \multicolumn{1}{c|}{15.01} & 12.14 \\
& M$^2$SNet \cite{ref_m2snet} & 2023 & \multicolumn{1}{c|}{80.23} & 3.80 & 6.66 & 9.19 & 10.08 & \multicolumn{1}{c|}{10.17} & 9.52 & \multicolumn{1}{c|}{12.33} & 9.66 \\
& CFANet \cite{ref_cfanet} & 2023 & \multicolumn{1}{c|}{79.27} & 1.69 & 6.44 & 9.90 & 10.44 & \multicolumn{1}{c|}{10.61} & 12.05 & \multicolumn{1}{c|}{14.29} & 10.62 \\ 
%\cline{2-12}

& CCLDNet-B \cite{ref_ccldnet} & 2022 & \multicolumn{1}{c|}{82.18} & 3.18 & 6.22 & 8.93 & 9.61 & \multicolumn{1}{c|}{12.19} & 10.77 & \multicolumn{1}{c|}{11.63} & 9.89 \\
& CCLDNet-L \cite{ref_ccldnet} & 2022 & \multicolumn{1}{c|}{82.75} & 2.39 & 4.58 & \textbf{7.58} & \textbf{7.63} & \multicolumn{1}{c|}{10.25} & 9.20 & \multicolumn{1}{c|}{\textbf{11.16}} & \textbf{8.40} \\
& Polyp-PVT \cite{ref_polyppvt} & 2022 & \multicolumn{1}{c|}{\textbf{83.68}} & 5.68 & 6.65 & 8.98 & 8.65 & \multicolumn{1}{c|}{\textbf{9.90}} & 10.29 & \multicolumn{1}{c|}{11.75} & 9.37 \\ \hline \hline

\multirow{3}{*}{\begin{sideways}Found.\end{sideways}} & Endo-FM \cite{ref_endofm} & 2023 & \multicolumn{1}{c|}{70.08} & 2.25 & 3.16 & 10.91 & 11.46 & \multicolumn{1}{c|}{12.93} & 11.91 & \multicolumn{1}{c|}{14.11} & 10.75 \\
& Polyp-SAM-B$^{*}$ \cite{ref_polypsam} & 2023 & \multicolumn{1}{c|}{93.16} & - & 0.75 & 2.66 & 2.89 & \multicolumn{1}{c|}{\textbf{2.69}} & \textbf{2.83} & \multicolumn{1}{c|}{2.52} & 2.39 \\
& Polyp-SAM-L$^{*}$ \cite{ref_polypsam} & 2023 & \multicolumn{1}{c|}{\textbf{93.23}} & - & \textbf{0.50} & \textbf{2.55} & \textbf{2.91} & \multicolumn{1}{c|}{2.85} & 2.90 & \multicolumn{1}{c|}{\textbf{2.22}} & \textbf{2.32} \\
& MedSAM$^{*}$ \cite{ref_medsam} & 2024 & \multicolumn{1}{c|}{90.48} & - & 1.13 & 3.63 & 3.65 & \multicolumn{1}{c|}{3.78} & 3.94 & \multicolumn{1}{c|}{3.52} & 3.28 \\
\hline \hline
\multicolumn{3}{c|}{Avg.} & \multicolumn{1}{c|}{81.20} & 2.81 & 4.68 & 8.83 & 9.34 & \multicolumn{1}{c|}{10.40} & 10.00 & \multicolumn{1}{c|}{11.70} & 9.16 \\
\bottomrule
\multicolumn{12}{l}{$^{\mathrm{a}}$ $*$ means that we use the bounding box of the polyp area as the input prompt.} \\
\multicolumn{12}{l}{$^{\mathrm{b}}$ Recon. means that we first add noises to the original images and then denoise them without any attribute transformation.}

\end{tabular}
}
\end{table*}

\begin{figure*}[h!]
    \centering
    \includegraphics[width=\linewidth]{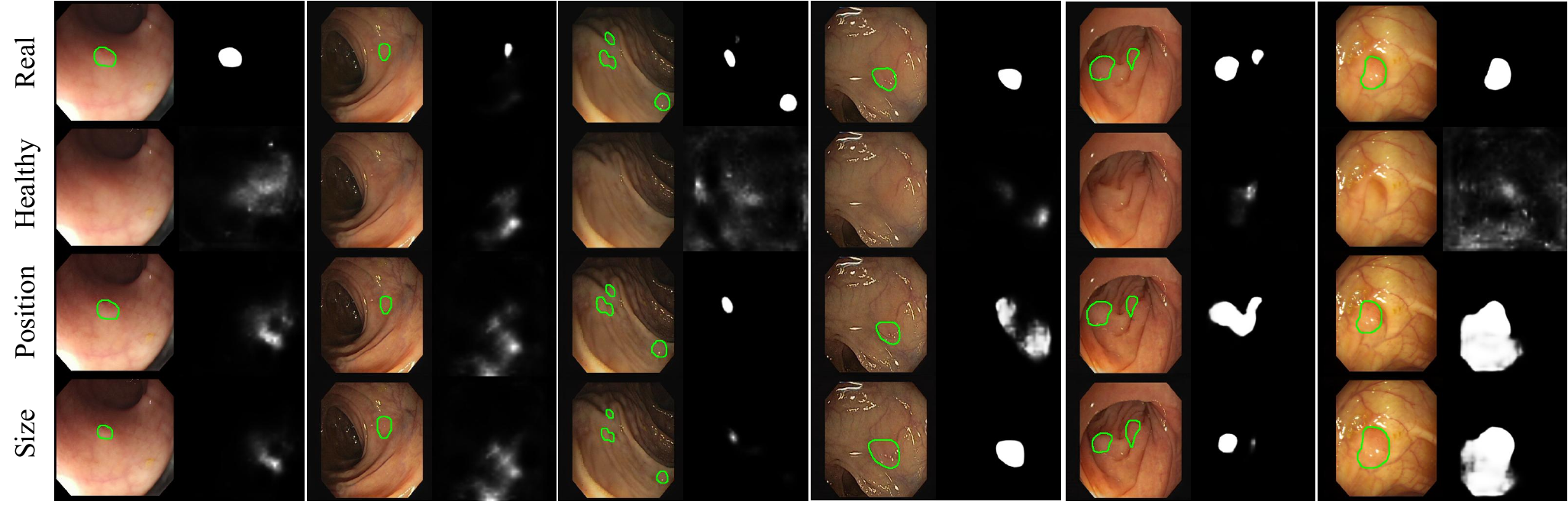}
    \caption{Example of segmentation results of PraNet ~\cite{ref_pranet} on the original dataset and proposed benchmark. The first column is the source images that outline the polyp boundary, the second column is the segmentation results.}
    \label{fig_segment}
\end{figure*}

The results are shown in Tab. \ref{tab_robustness}, and we report the average \textit{Dice} drop under all polyp size and position changes in the last column.
We have the following observations.

\begin{enumerate}
    \item Our edited images, featuring polyp existence, size, and position attribute changes, reduce performance across all models to varying extents. 
    \item In pure healthy samples, all methods' results exhibit varying degrees of False Positive Rate. This inspires that segmentation models should consider healthy samples as applied in clinical scenarios.
    \item The attribute robustness of models decreases with increasing attribute variations. For example, the average performance drop arises from 9.20\% to 10.87\% as the size change magnitude increases from 0.1 to 0.3. 
    \item The robustness under attribute changes is improved along with improvements in original segmentation performances on different models, such as switching from the PraNet \cite{ref_pranet} to the Polyp-PVT \cite{ref_polyppvt}. By this measure, models have become more and more capable of generalizing to different clinical scenarios, which implies that they indeed learn some robust features. This implies that attribute robustness can be a practical manner to evaluate future progress in representation learning. 
    \item The recent foundation model Polyp-SAM \cite{ref_polypsam} and MedSAM \cite{ref_medsam}, which uses the Segment-Anything-Model (SAM) \cite{ref_sam} as the backbone, shows the best performances in both real and edited datasets. Qualitative segmentation results are shown in Fig. \ref{fig_segment}.
\end{enumerate}

%As shown in Tab. \ref{tab_robustness}, when only the size or position of the polyp is edited, the Dice score decreases to different degrees, leading to model performance degradation. We also find that the robustness under different attributes is improved along with improvements in terms of the original Dice score on different models, such as switching from PraNet \cite{ref_pranet} to Polyp-PVT \cite{ref_polyppvt} method. By this measure, models have become more and more capable of generalizing to different clinical scenarios, which implies that they indeed learn some robust features. It shows that attribute robustness can be a good way to measure future progress in representation learning. We also observe that larger networks possess better robustness on the attribute editing, the overall robustness of CNN networks is lower than Transformer-based networks, while Transformer-based networks exhibit inferior robustness compared to the Foundation models. Consequently, models with even more depth, width, and feature aggregation may attain further attribute robustness. We also observed that while recent models have shown significant improvements in polyp segmentation, they also exhibit more instances of erroneous segmentation in healthy images.

\begin{table*}[h!]
    \centering
    \footnotesize
    \caption{The quantitative results using different augmentation techniques.}
    \renewcommand{\arraystretch}{1.2} % 行距
    \setlength{\tabcolsep}{6pt} % 像素列宽，默认6像素
    \begin{tabular}{r|ccc|ccc}
    \toprule
    & \multicolumn{3}{c|}{In-Distribution} & \multicolumn{3}{c}{Out-of-Distribution} \\
    \cline{2-7}
    & PraNet \cite{ref_pranet} & SANet \cite{ref_sanet} & PVT \cite{ref_polyppvt} & PraNet \cite{ref_pranet} & SANet \cite{ref_sanet} & PVT \cite{ref_polyppvt}    \\
    \hline
    \textit{w/o augmentation} & 74.5   & 71.4  & 76.0  & 75.3 & 80.3 & 81.4 \\
    \hline
    SDM \cite{ref_sdm}   & +1.9   & +1.5  & +0.3  &  -1.0  & -1.6 & -2.7 \\
    ArSDM \cite{ref_arsdm}   & +5.5   & +2.7  & +0.6  &  -0.6  & -1.6 & -3.8  \\
    ImageNet-E \cite{ref_imagenete}   & +6.6 & +6.4 & +6.2 & -1.4 & -4.2 & -2.2 \\
    \rowcolor{mygray} \textbf{Ours}  & \textbf{+7.9}   & \textbf{+8.2}  & \textbf{+6.6} & \textbf{+0.8} &\textbf{+1.8}&\textbf{+0.5}\\
    \bottomrule
    \end{tabular}
    \label{tab_augmentation}
\end{table*}

\subsection{Practical Application} 
To demonstrate the effectiveness of our proposed pipeline on improving segmentation robustness, we compare our methods with previous image synthesis methods, including SDM \cite{ref_sdm}, ArSDM \cite{ref_arsdm} and ImageNet-E \cite{ref_imagenete}, by adopting them as data augmentation techniques in training. 
We construct the polyp size and position attribute edited training sets based on data split configuration of \cite{ref_pranet}, and mix them with real data to train segmentation models PraNet \cite{ref_pranet}, SANet \cite{ref_sanet} and Polyp-PVT \cite{ref_polyppvt}. We evaluate performances on In-Distribution data (combination of five test sets) and Out-of-Distribution (OOD) PICCOLO dataset \cite{ref_piccolo}. % and our curated Polyp-E 

%so we apply the proposed method to perform size editing and position editing on the image-mask pairs from the real training dataset, generating an equal number of samples. Subsequently, these augmented samples were combined with the original training dataset to form a novel downstream training set. We employed PraNet \cite{ref_pranet} as the baseline segmentation model under default settings and evaluated it alongside various data augmentation methods.

The results are shown in Tab. \ref{tab_augmentation}. Although SDM \cite{ref_sdm} and ImageNet-E\cite{ref_imagenete} were not specifically designed for colonoscopy images and may synthesize samples that lack the ability for effective improvement, the results verify the superiority of our approach.
It is noteworthy that, using our generated attribute variations can improve models' generalization ability under the OOD dataset while others impair models' generalization ability. For instance, using ArSDM \cite{ref_arsdm}, the result of SANet \cite{ref_sanet} declines 1.6\%; but using our samples, it raises 1.8\% when evaluating in the OOD dataset. This observation indicates considering attribute changes is significant in improving models' capability towards more complex scenarios.

%Subsequently, the results in our Polyp-E, show that our pipeline as a form of data augmentation significantly reduces segmentation models' sensitivity to variations in polyp attributes. This study also implies the significance of considering polyp attribute variations in clinical applications.

\section{Conclusion}
In this paper, we propose a diffusion-based image editing pipeline that generates faithful polyp attribute variations. Using the proposed pipeline, we construct the Polyp-E dataset to benchmark robustness under different object attribute variations including healthy, polyp size, and positions. Extensive experiments on different state-of-the-art polyp segmentation models exhibit varying sensitivities to attribute changes. Moreover, recent SAM-based foundation methods show greater results than conventional models. We further investigate the superiority of our pipeline in enhancing models' robustness and generalization ability by comparing it with other popular image synthesis approaches.

% \\ \hspace*{\fill} \\

\bibliographystyle{IEEEtran}
\bibliography{mybibliography}

\begin{thebibliography}{10}
\providecommand{\url}[1]{#1}
\csname url@samestyle\endcsname
\providecommand{\newblock}{\relax}
\providecommand{\bibinfo}[2]{#2}
\providecommand{\BIBentrySTDinterwordspacing}{\spaceskip=0pt\relax}
\providecommand{\BIBentryALTinterwordstretchfactor}{4}
\providecommand{\BIBentryALTinterwordspacing}{\spaceskip=\fontdimen2\font plus
\BIBentryALTinterwordstretchfactor\fontdimen3\font minus \fontdimen4\font\relax}
\providecommand{\BIBforeignlanguage}[2]{{%
\expandafter\ifx\csname l@#1\endcsname\relax
\typeout{** WARNING: IEEEtran.bst: No hyphenation pattern has been}%
\typeout{** loaded for the language `#1'. Using the pattern for}%
\typeout{** the default language instead.}%
\else
\language=\csname l@#1\endcsname
\fi
#2}}
\providecommand{\BIBdecl}{\relax}
\BIBdecl

\bibitem{siegel2020colorectal}
R.~L. Siegel, K.~D. Miller, A.~Goding~Sauer, S.~A. Fedewa, L.~F. Butterly, J.~C. Anderson, A.~Cercek, R.~A. Smith, and A.~Jemal, ``Colorectal cancer statistics, 2020,'' \emph{CA: a cancer journal for clinicians}, vol.~70, no.~3, pp. 145--164, 2020.

\bibitem{ali2024assessing}
S.~Ali, N.~Ghatwary, D.~Jha, E.~Isik-Polat, G.~Polat, C.~Yang, W.~Li, A.~Galdran, M.-{\'A}.~G. Ballester, V.~Thambawita \emph{et~al.}, ``Assessing generalisability of deep learning-based polyp detection and segmentation methods through a computer vision challenge,'' \emph{Scientific Reports}, vol.~14, no.~1, p. 2032, 2024.

\bibitem{ref_unet}
O.~Ronneberger, P.~Fischer, and T.~Brox, ``U-net: Convolutional networks for biomedical image segmentation,'' in \emph{Medical Image Computing and Computer-Assisted Intervention--MICCAI 2015: 18th International Conference, Munich, Germany, October 5-9, 2015, Proceedings, Part III 18}.\hskip 1em plus 0.5em minus 0.4em\relax Springer, 2015, pp. 234--241.

\bibitem{ref_pranet}
D.-P. Fan, G.-P. Ji, T.~Zhou, G.~Chen, H.~Fu, J.~Shen, and L.~Shao, ``Pranet: Parallel reverse attention network for polyp segmentation,'' in \emph{International conference on medical image computing and computer-assisted intervention}.\hskip 1em plus 0.5em minus 0.4em\relax Springer, 2020, pp. 263--273.

\bibitem{ref_uacanet}
T.~Kim, H.~Lee, and D.~Kim, ``Uacanet: Uncertainty augmented context attention for polyp segmentation,'' in \emph{Proceedings of the 29th ACM International Conference on Multimedia}, 2021, pp. 2167--2175.

\bibitem{ref_sanet}
J.~Wei, Y.~Hu, R.~Zhang, Z.~Li, S.~K. Zhou, and S.~Cui, ``Shallow attention network for polyp segmentation,'' in \emph{Medical Image Computing and Computer Assisted Intervention--MICCAI 2021: 24th International Conference, Strasbourg, France, September 27--October 1, 2021, Proceedings, Part I 24}.\hskip 1em plus 0.5em minus 0.4em\relax Springer, 2021, pp. 699--708.

\bibitem{ref_msnet}
Q.~Liu, Q.~Dou, L.~Yu, and P.~A. Heng, ``Ms-net: multi-site network for improving prostate segmentation with heterogeneous mri data,'' \emph{IEEE transactions on medical imaging}, vol.~39, no.~9, pp. 2713--2724, 2020.

\bibitem{ref_m2snet}
X.~Zhao, H.~Jia, Y.~Pang, L.~Lv, F.~Tian, L.~Zhang, W.~Sun, and H.~Lu, ``M$^2$snet: Multi-scale in multi-scale subtraction network for medical image segmentation,'' \emph{arXiv preprint arXiv:2303.10894}, 2023.

\bibitem{ref_cfanet}
T.~Zhou, Y.~Zhou, K.~He, C.~Gong, J.~Yang, H.~Fu, and D.~Shen, ``Cross-level feature aggregation network for polyp segmentation,'' \emph{Pattern Recognition}, vol. 140, p. 109555, 2023.

\bibitem{ref_ccldnet}
H.~Yang, Q.~Chen, K.~Fu, L.~Zhu, L.~Jin, B.~Qiu, Q.~Ren, H.~Du, and Y.~Lu, ``Boosting medical image segmentation via conditional-synergistic convolution and lesion decoupling,'' \emph{Computerized Medical Imaging and Graphics}, vol. 101, p. 102110, 2022.

\bibitem{xiao2022icbnet}
Y.~Xiao, Z.~Chen, L.~Wan, L.~Yu, and L.~Zhu, ``Icbnet: Iterative context-boundary feedback network for polyp segmentation,'' in \emph{2022 IEEE International Conference on Bioinformatics and Biomedicine (BIBM)}.\hskip 1em plus 0.5em minus 0.4em\relax IEEE, 2022, pp. 1297--1304.

\bibitem{ref_polyppvt}
B.~Dong, W.~Wang, D.-P. Fan, J.~Li, H.~Fu, and L.~Shao, ``Polyp-pvt: Polyp segmentation with pyramid vision transformers,'' \emph{arXiv preprint arXiv:2108.06932}, 2021.

\bibitem{huang2022transmixer}
Y.~Huang, D.~Tan, Y.~Zhang, X.~Li, and K.~Hu, ``Transmixer: A hybrid transformer and cnn architecture for polyp segmentation,'' in \emph{2022 IEEE International Conference on Bioinformatics and Biomedicine (BIBM)}.\hskip 1em plus 0.5em minus 0.4em\relax IEEE, 2022, pp. 1558--1561.

\bibitem{ref_endofm}
Z.~Wang, C.~Liu, S.~Zhang, and Q.~Dou, ``Foundation model for endoscopy video analysis via large-scale self-supervised pre-train,'' in \emph{International Conference on Medical Image Computing and Computer-Assisted Intervention}.\hskip 1em plus 0.5em minus 0.4em\relax Springer, 2023, pp. 101--111.

\bibitem{ref_polypsam}
Y.~Li, M.~Hu, and X.~Yang, ``Polyp-sam: Transfer sam for polyp segmentation,'' \emph{arXiv preprint arXiv:2305.00293}, 2023.

\bibitem{ref_medsam}
J.~Ma, Y.~He, F.~Li, L.~Han, C.~You, and B.~Wang, ``Segment anything in medical images,'' \emph{Nature Communications}, vol.~15, no.~1, p. 654, 2024.

\bibitem{ref_ppsam}
M.~M. Rahman, M.~Munir, D.~Jha, U.~Bagci, and R.~Marculescu, ``Pp-sam: Perturbed prompts for robust adaptation of segment anything model for polyp segmentation,'' \emph{arXiv preprint arXiv:2405.16740}, 2024.

\bibitem{xia2024dusformer}
Z.~Xia, J.~Chen, and C.~Lu, ``Dusformer: Dual-swin transformer v2 aggregate network for polyp segmentation,'' \emph{IEEE Access}, 2024.

\bibitem{ref_sunseg}
G.-P. Ji, G.~Xiao, Y.-C. Chou, D.-P. Fan, K.~Zhao, G.~Chen, and L.~Van~Gool, ``Video polyp segmentation: A deep learning perspective,'' \emph{Machine Intelligence Research}, vol.~19, no.~6, pp. 531--549, 2022.

\bibitem{ref_piccolo}
L.~F. S{\'a}nchez-Peralta, J.~B. Pagador, A.~Pic{\'o}n, {\'A}.~J. Calder{\'o}n, F.~Polo, N.~Andraka, R.~Bilbao, B.~Glover, C.~L. Saratxaga, and F.~M. S{\'a}nchez-Margallo, ``Piccolo white-light and narrow-band imaging colonoscopic dataset: A performance comparative of models and datasets,'' \emph{Applied Sciences}, vol.~10, no.~23, p. 8501, 2020.

\bibitem{ref_kvasir}
D.~Jha, P.~H. Smedsrud, M.~A. Riegler, P.~Halvorsen, T.~de~Lange, D.~Johansen, and H.~D. Johansen, ``Kvasir-seg: A segmented polyp dataset,'' in \emph{MultiMedia Modeling: 26th International Conference, MMM 2020, Daejeon, South Korea, January 5--8, 2020, Proceedings, Part II 26}.\hskip 1em plus 0.5em minus 0.4em\relax Springer, 2020, pp. 451--462.

\bibitem{ref_ddpm}
J.~Ho, A.~Jain, and P.~Abbeel, ``Denoising diffusion probabilistic models,'' \emph{Advances in neural information processing systems}, vol.~33, pp. 6840--6851, 2020.

\bibitem{ref_ddim}
J.~Song, C.~Meng, and S.~Ermon, ``Denoising diffusion implicit models,'' \emph{arXiv preprint arXiv:2010.02502}, 2020.

\bibitem{ref_ldm}
R.~Rombach, A.~Blattmann, D.~Lorenz, P.~Esser, and B.~Ommer, ``High-resolution image synthesis with latent diffusion models,'' in \emph{Proceedings of the IEEE/CVF conference on computer vision and pattern recognition}, 2022, pp. 10\,684--10\,695.

\bibitem{ref_sdm}
W.~Wang, J.~Bao, W.~Zhou, D.~Chen, D.~Chen, L.~Yuan, and H.~Li, ``Semantic image synthesis via diffusion models,'' 2022.

\bibitem{ref_arsdm}
Y.~Du, Y.~Jiang, S.~Tan, X.~Wu, Q.~Dou, Z.~Li, G.~Li, and X.~Wan, ``Arsdm: colonoscopy images synthesis with adaptive refinement semantic diffusion models,'' in \emph{International conference on medical image computing and computer-assisted intervention}.\hskip 1em plus 0.5em minus 0.4em\relax Springer, 2023, pp. 339--349.

\bibitem{ref_imagenete}
X.~Li, Y.~Chen, Y.~Zhu, S.~Wang, R.~Zhang, and H.~Xue, ``Imagenet-e: Benchmarking neural network robustness via attribute editing,'' in \emph{Proceedings of the IEEE/CVF Conference on Computer Vision and Pattern Recognition}, 2023, pp. 20\,371--20\,381.

\bibitem{liao2020exploration}
R.~Liao, K.~Qi, D.~Che, and T.~H. Zeng, ``Exploration of the possibility of early diagnosis for digestive diseases using deep learning techniques,'' in \emph{2020 IEEE International Conference on Bioinformatics and Biomedicine (BIBM)}.\hskip 1em plus 0.5em minus 0.4em\relax IEEE, 2020, pp. 2343--2350.

\bibitem{chen2018reverse}
S.~Chen, X.~Tan, B.~Wang, and X.~Hu, ``Reverse attention for salient object detection,'' in \emph{Proceedings of the European conference on computer vision (ECCV)}, 2018, pp. 234--250.

\bibitem{min2022transformer}
E.~Min, R.~Chen, Y.~Bian, T.~Xu, K.~Zhao, W.~Huang, P.~Zhao, J.~Huang, S.~Ananiadou, and Y.~Rong, ``Transformer for graphs: An overview from architecture perspective,'' \emph{arXiv preprint arXiv:2202.08455}, 2022.

\bibitem{ref_sam}
A.~Kirillov, E.~Mintun, N.~Ravi, H.~Mao, C.~Rolland, L.~Gustafson, T.~Xiao, S.~Whitehead, A.~C. Berg, W.-Y. Lo \emph{et~al.}, ``Segment anything,'' \emph{arXiv preprint arXiv:2304.02643}, 2023.

\bibitem{hendrycks2019benchmarking}
D.~Hendrycks and T.~Dietterich, ``Benchmarking neural network robustness to common corruptions and perturbations,'' \emph{arXiv preprint arXiv:1903.12261}, 2019.

\bibitem{hendrycks2019using}
D.~Hendrycks, K.~Lee, and M.~Mazeika, ``Using pre-training can improve model robustness and uncertainty,'' in \emph{International conference on machine learning}.\hskip 1em plus 0.5em minus 0.4em\relax PMLR, 2019, pp. 2712--2721.

\bibitem{ali2022endoscopic}
S.~Ali and N.~M. Ghatwary, ``Endoscopic computer vision challenges 2.0.'' in \emph{EndoCV@ ISBI}, 2022, pp. 5--8.

\bibitem{eche2021toward}
T.~Eche, L.~H. Schwartz, F.-Z. Mokrane, and L.~Dercle, ``Toward generalizability in the deployment of artificial intelligence in radiology: role of computation stress testing to overcome underspecification,'' \emph{Radiology: Artificial Intelligence}, vol.~3, no.~6, p. e210097, 2021.

\bibitem{boone2023rood}
L.~Boone, M.~Biparva, P.~M. Forooshani, J.~Ramirez, M.~Masellis, R.~Bartha, S.~Symons, S.~Strother, S.~E. Black, C.~Heyn \emph{et~al.}, ``Rood-mri: Benchmarking the robustness of deep learning segmentation models to out-of-distribution and corrupted data in mri,'' \emph{NeuroImage}, vol. 278, p. 120289, 2023.

\bibitem{maron2021benchmark}
R.~C. Maron, J.~G. Schlager, S.~Haggenm{\"u}ller, C.~von Kalle, J.~S. Utikal, F.~Meier, F.~F. Gellrich, S.~Hobelsberger, A.~Hauschild, L.~French \emph{et~al.}, ``A benchmark for neural network robustness in skin cancer classification,'' \emph{European Journal of Cancer}, vol. 155, pp. 191--199, 2021.

\bibitem{young2021stress}
A.~T. Young, K.~Fernandez, J.~Pfau, R.~Reddy, N.~A. Cao, M.~Y. von Franque, A.~Johal, B.~V. Wu, R.~R. Wu, J.~Y. Chen \emph{et~al.}, ``Stress testing reveals gaps in clinic readiness of image-based diagnostic artificial intelligence models,'' \emph{NPJ digital medicine}, vol.~4, no.~1, p.~10, 2021.

\bibitem{islam2023robustness}
M.~Islam, Z.~Li, and B.~Glocker, ``Robustness stress testing in medical image classification,'' in \emph{International Workshop on Uncertainty for Safe Utilization of Machine Learning in Medical Imaging}.\hskip 1em plus 0.5em minus 0.4em\relax Springer, 2023, pp. 167--176.

\bibitem{ref_nonequilibrium}
J.~Sohl-Dickstein, E.~Weiss, N.~Maheswaranathan, and S.~Ganguli, ``Deep unsupervised learning using nonequilibrium thermodynamics,'' in \emph{International conference on machine learning}.\hskip 1em plus 0.5em minus 0.4em\relax PMLR, 2015, pp. 2256--2265.

\bibitem{ref_vae}
D.~P. Kingma and M.~Welling, ``Auto-encoding variational bayes,'' \emph{arXiv preprint arXiv:1312.6114}, 2013.

\bibitem{ref_attack}
A.~Arnab, O.~Miksik, and P.~H. Torr, ``On the robustness of semantic segmentation models to adversarial attacks,'' in \emph{Proceedings of the IEEE conference on computer vision and pattern recognition}, 2018, pp. 888--897.

\bibitem{ref_endoscene}
D.~V{\'a}zquez, J.~Bernal, F.~J. S{\'a}nchez, G.~Fern{\'a}ndez-Esparrach, A.~M. L{\'o}pez, A.~Romero, M.~Drozdzal, A.~Courville \emph{et~al.}, ``A benchmark for endoluminal scene segmentation of colonoscopy images,'' \emph{Journal of healthcare engineering}, vol. 2017, 2017.

\bibitem{ref_clinicdb}
J.~Bernal, F.~J. S{\'a}nchez, G.~Fern{\'a}ndez-Esparrach, D.~Gil, C.~Rodr{\'\i}guez, and F.~Vilari{\~n}o, ``Wm-dova maps for accurate polyp highlighting in colonoscopy: Validation vs. saliency maps from physicians,'' \emph{Computerized medical imaging and graphics}, vol.~43, pp. 99--111, 2015.

\bibitem{ref_colondb}
N.~Tajbakhsh, S.~R. Gurudu, and J.~Liang, ``Automated polyp detection in colonoscopy videos using shape and context information,'' \emph{IEEE transactions on medical imaging}, vol.~35, no.~2, pp. 630--644, 2015.

\bibitem{ref_etis}
J.~Silva, A.~Histace, O.~Romain, X.~Dray, and B.~Granado, ``Toward embedded detection of polyps in wce images for early diagnosis of colorectal cancer,'' \emph{International journal of computer assisted radiology and surgery}, vol.~9, pp. 283--293, 2014.

\bibitem{ref_fid}
M.~Heusel, H.~Ramsauer, T.~Unterthiner, B.~Nessler, and S.~Hochreiter, ``Gans trained by a two time-scale update rule converge to a local nash equilibrium,'' \emph{Advances in neural information processing systems}, vol.~30, 2017.

\end{thebibliography}

\end{document}